\documentclass{article}

\usepackage{arxiv}
\usepackage[utf8]{inputenc} 
\usepackage[T1]{fontenc}    
\usepackage{url}        
\usepackage{booktabs}       
\usepackage{amsfonts}   
\usepackage{nicefrac}   
\usepackage{longtable}
\usepackage{microtype}      
\usepackage{lipsum}
\usepackage{graphicx}
\usepackage[backend=bibtex]{biblatex}
\bibliography{references} 
\begin{document}
\title{cooking is all about people: comment classification on cookery channels
using bert and classification models (malayalam-english mix-code)
}

\author{
 Subramaniam Kazhuparambil  \\
  School of Computing\\
  Dublin Business School\\
  D02 WC04 Dublin, Ireland \\
  \texttt{10524303@mydbs.ie} \\
\and
\textbf{Abhishek Kaushik} \\
  ADAPT Centre, School of Computing\\
  Dublin City University\\
  D09 W6Y4 Dublin, Ireland \\
  \texttt{abhishek.kaushik2@mail.dcu.ie} \\
}

\date{}
\maketitle

\section*{Abstract}
\textbf{The scope of a lucrative career promoted by Google through its video distribution platform YouTube \footnote{https://www.youtube.com/} has attracted a large number of users to become content creators. An important aspect of this line of work is the feedback received in the form of comments which show how well the content is being received by the audience. However, volume of comments coupled with spam and limited tools for comment classification makes it virtually impossible for a creator to go through each and every comment and gather constructive feedback. Automatic classification of comments is a challenge even for established classification models, since comments are often of variable lengths riddled with slang, symbols and abbreviations. This is a greater challenge where comments are multilingual as the messages are often rife with the respective vernacular. In this work, we have evaluated top-performing classification models for classifying comments which are a mix of different combinations of English and Malayalam (only English, only Malayalam and Mix of English and Malayalam). The statistical analysis of results indicates that Multinomial Naïve Bayes, K-Nearest Neighbors (KNN), Support Vector Machine (SVM), Random Forest and Decision Trees offer similar level of accuracy in comment classification. Further, we have also evaluated 3 multilingual transformer based language models (BERT, DISTILBERT and XLM) and compared their performance to the traditional machine learning classification techniques. XLM was the top-performing model with an accuracy of 67.31\%. Random Forest with Term Frequency Vectorizer was the best performing model out of all the traditional classification models with an accuracy of 63.59\%.}

{\textit{\textbf{Keywords - BERT, Classification, Mix-Code, Language Model, Youtube, Parametric and Non-Parametric.}}} 
\\
\section{Introduction}
The dissemination of broadband internet around the world has resulted into a massive influx of Internet users.  The accessibility coupled with faster internet connection has made video streaming and hosting services extremely popular amongst the users. According to an article of India Times \footnote{https://timesofindia.indiatimes.com/}, video consumption, especially through YouTube and Netflix \footnote{https://www.netflix.com/}, accounts for 58\% of the total internet traffic. The availability of internet services and resources, accessible because of reliable broadband connection allowed the inception of new and sophisticated web-based platforms. YouTube was one of the newly conceived platforms in the era of Web 2.0. YouTube allowed creators to publish video content. Moreover, it incorporated social networking features, such as the ability to post comments. The comment section acted as a medium of interaction between the content creator and viewers. YouTube has emerged to become the global standard for video consumption and its success is evident in the form of recent statistics reported by Google: The platform has over 2 billion users; over 500 hours of video content is uploaded every minute; and over one billion hours of content is consumed daily. YouTube has launched local versions in more than 100 countries supporting over 80 different languages. More than 70\% of the total watch time comes from mobile devices and YouTube accounts for 37\% of the entire mobile internet traffic.
In recent years, YouTube has adopted a monetization model to reward content creators, stimulating them to produce high quality original video content. The monetary aspect of creating YouTube videos has made viewer feedback more relevant than ever as the revenue generated from a video is highly correlated with the feedback. Hence, it is imperative for a YouTube producer to regularly go through viewer feedback and use this insight to improve video quality.
For the producer of a successful YouTube channel, manually skimming through comments is infeasible due to the large volume. A text-classification model that can automatically identify the emotion of a comment and classify it into predefined labels can facilitate seamless evaluation of viewer response \cite{shah2019sentiment, venkatakrishnan2020sentiment, kaushikstudy}.
Established models for text-classification have encountered challenges when dealing with YouTube comments. This is mainly due to the fact that these comments are often of variable length and rife with slang, symbols, emoticons and abbreviations. More importantly, YouTube supporting over 80 languages adds another aspect to consider when classifying multilingual comments. This makes tokenization of comments a challenging task.\\\\
Given this scenario, this paper presents a comprehensive performance evaluation of top-performing machine learning classification models that can be applied to automatically classify comments. Further, performance of the established techniques is compared to the novel transformer based language models: BERT \footnote{https://arxiv.org/pdf/1706.03762.pdf} (Bidirectional Encoder Representations from Transformers), DISTILBERT and XLM (DISTILBERT AND XLM are extensions of BERT). BERT was the result of a recent paper published by researchers at Google AI Language. The main goal of this paper is to find promising techniques and settings which can be translated into a tool to facilitate text-classification of YouTube comments. Additionally, this paper aims to answer the following research questions:
\begin{itemize}
    \item How difficult it is to classify multilingual text data?
    \item What will be the statistical significance of the results obtained from multiple classification models?
\end{itemize}
The remainder of this paper is structured as follows: Section 2 describes the related work available in the literature. In Section 3, we present the dataset description, experimental methodology and configurations used to perform the experiment. Section 4 elaborates the achieved results. Section 5 sheds light on the challenges faced and discusses implications of the achieved results. Finally, Section 6 describes the main conclusion and offers suggestions for future work.
\\
\section{Literature Review}

This section is a review of studies related to text classification. In this section, we discuss and compare some of the existing research; applied techniques; limitations and computational difficulties; and model performances. To write this review, we investigated around 60-70 papers, out of which 30 papers were selected based on diversity of application and models used. The review is divided into six sections: \textit{Text Processing}, \textit{Machine Learning}, \textit{Deep Learning Models}, \textit{Hybrid Models} and \textit{BERT}. Table \ref{table1} presents the literature survey of the selected papers.

\begin{longtable}{p{3cm}p{4.3cm}p{4.3cm}p{3cm}}
\caption{\textbf{Literature survey table of selected works}}
\label{table1}
\\
\hline \\ 
     \textbf{Authors} &  \textbf{Datasets} &  \textbf{Methods Uzsed} & \textbf{Results} \\\\ \hline \\
        %1&
        Chanchana et al. \cite{sornsoontorn2017using}& Thesis papers and journal of Kasetsart University.& LR, Naive Bayes(NB), SVM,DT, RF, Ada Boost (ADB) and KNN & 92.7 (LR, NB, SVM, DT, RF, ADB and KNN)
        \\\\ \hline \\
        Aman and Stan. \cite{aman2007identifying}& Blog posts data collected from the internet.& SVM and NB & 73.89(SVM and NB)
        \\\\ \hline \\
        Anton. \cite{Anthon2017text}& Swedish language dataset, and English language dataset & SVM, LR and Neural Networks & 0.631 F1-score (SVM, LR and Neural Networks)
        \\\\ \hline \\
        Durant and Smith. \cite{durant2006mining} & Politial weblog from website http://www.themoderatevoice.com & SVM and NB & 78.06 (SVM and NB)
        \\\\ \hline \\
        Wang, Julia and Tang. \cite{wang2003classification}&722 documents downloaded from the Web & NB-B, NB-M & 67.19 (NB-B,NB-M)
        \\\\ \hline \\
        Joshi Kalyani et al. \cite{kalyani2016stock}&Apple Inc. Company’s data from 2013 to 2016 & SVM, NB and RF & 92 (SVM, Naive Bayes and RF)
        \\\\ \hline \\
        Chaitanya Anne. \cite{anne2017advanced} & NASA’s patent documents by U.S. Patent Office & SVM, KNN, 2 tree RF and J48 & 96.03 (SVM, KNN, 2 tree RF and J48)
        \\\\ \hline \\
        Kleverwal. \cite{kleverwal2015supervised}& unstructured texts and related complaints & KNN & 0.731 (OR-query  and  Unigrams with KNN)
        \\\\ \hline \\
        T Crijns. \cite{crijns2016text}&SQL Database of Ugenda&SVM, LR, NB and RF & 0.8110 F1-score (SVM, LR, NB and RF)\\
        \\\\ \hline \\
        Tobias \cite{tobias2017automatic}&Data from real web pages provided by Whaam.& SVM, KNN, NB-M, TWCNB and N-Gram & 70 (SVM,KNN,NB-M, TWCNB and N-Gram)
        \\\\ \hline \\
        Heidarysafa et al. \cite{heidarysafa2018analysis}&Federal Railroad Administration (FRA) reports.& CNN, RNN and DNN &75 (CNN,RNN and DNN)
        \\\\ \hline \\
        Lin Li et al. \cite{li2017text} & NetEase news data and text classification & CNN & 96.97 (CNN)
        \\\\ \hline \\
        F Grawe et al. \cite{grawe2017automated}& USPTO Bulk Data from (https://bulkdata.uspto.gov/).& LSTM & 55 (SVM , LSTM)
        \\\\ \hline \\
        Bo Shu et al. \cite{shu2018investigating}& DBPendia, Yelp Review Polarity, Yahoo! Answers, Amazon Reviews & CNN and LSTM & Prospective future work (CNN and LSTM)
        \\\\ \hline \\
        Zharmagambetov et al. \cite{zharmagambetov2015sentiment}& IMDB  movie reviews.& RNN & 89.8 (RNN)
        \\\\ \hline \\
        Qinlu Zhao et al. \cite{zhao2017commented} & COAE2014 task4 weibo dataset.& CNN with attention mechanism & 95.15 (CNN with attention mechanism)
        \\\\ \hline \\
        Shamik Kundu el al. \cite{kundu2018classification} & Forum for Information Retrieval Evaluation (FIRE).& LSTM & 0.9159 F1-score (LSTM)
        \\\\ \hline \\
        Dehua Chen et al. \cite{chen2017deep}&Real medical data and the new-thyroid.& RF, SVM and Deep Neural Network & 95 (RF, SVM and Deep Neural Network)
        \\\\ \hline \\
        Suyash and Xavier. \cite{lakhotia2018experimental}&Rotten Tomatoes Sentence Polarity, 20 Newsgroups and Reuters Corpus.& SVM, MLP to CNN and with graph theory & 91.44 (SVM, MLP to CNN and with graph theory)
        \\\\ \hline \\
        Afakh et al. \cite{afakh2017aksara}&CDAR 2003 training set, Aksara Jawa images & CNN & 0.96 Precision (CNN)
        \\\\ \hline \\
        Mitra and Reza. \cite{behzadi2018text}& 200,000 instances (SynthText) and ICDAR dataset with 7200 images & Fully Convolutional DenseNets & 0.93 (CNN, Fully Convolutional DenseNets)
        \\\\ \hline \\
        Khaleel and Buket. \cite{ameen2018spam} & 5-days tweet data & MLP with Naive Bayes, RF and J48 & 92 (MLP with Naive Bayes, RF and J48)
        \\\\ \hline \\
        Lan-Juan Li et al. \cite{li2018electronic} & Medical record files of two top-3 hospitals.& SVM, LSTM and Yoon model & 94.7 (SVM, LSTM and Yoon model.)
        \\\\ \hline \\
        Alper and Murphey. \cite{uysal2017sentiment} & IMDB movie review, subset of Sentiment 140 and Nine Public Sentiment dataset & CNN+LSTM & 85.4 (CNN+LSTM)
        \\\\ \hline \\
        Thazhackal and Susheela. \cite{thazhackal2018hybrid}& Bookstores dataset which is present in the Yelp dataset & DT, SVM, RF, GBoost, CNN and DNN with transfer learning & 89.3 (DT, SVM, RF, GBoost, CNN and DNN with transfer learning)
        \\\\ \hline \\
        Phaisangittisagul et al. \cite{phaisangittisagul2019target}&Data from promotional advertising at www.wongnai.com.&Deep Learning&90.15(deep learning)
        \\\\ \hline \\
        Sreekanth and Sankar. \cite{madisetty2018neural}& HSpam14, 1KS10KN dataset & CNN & 95(CNN,RF,Ensemmble)
        \\\\ \hline \\
        Anupama Ray et al. \cite{ray2015text}&Printed  oriya  text.&Deep  Bidirectional LSTM & 96.3 (Deep  Bidirectional LSTM)
        \\\\ \hline \\
        Wang and Bohao. \cite{wang2019research}& Data from sohu website crawl. & double Bi-Gated Recurrent Unit & 0.805 F1-Score (double Bi-Gated Recurrent Unit)
        \\\\ \hline \\
        Xianglong Chen et al. \cite{chen2018hybrid} & NLPCC2018  task  6.& BiGRU and CNN & 44.7 F1-score (BiGRU+CNN)
        \\\\ \hline 
\end{longtable}

\subsection{Text Processing}

\subsubsection{Analysis of Railway Accidents' Narratives using Deep Learning}

This paper \cite{heidarysafa2018analysis} states, from 2001 to 2016, rail accidents in the U.S. cost more than 4.6 Billion Dollars. The accident details reported to the Federal Railroad Administration (FRA) consists of entries such as primary cause and a short description. It is not easy for a non-expert reader to understand the terminologies. Therefore, developing a model to label these accidents based on primary cause would help us understand the narrative and we can find out the consistent nature of the accidents. This paper experiments with data extracted from accidents which have occurred in the past. It applies Deep Learning combined with word embedding (such as Word2Vec and GloVe) to classify accidents using text in the narratives . The data used are the reports submitted to FRA. The Deep Learning techniques used for this experiment are Convolutional Neural Networks (CNN), Recurrent Neural Networks (RNN), and Deep Neural Networks (DNN). The results obtained from the Deep Learning models are compared with that of traditional machine Learning algorithms such as Support Vector Machines (SVM), Naive Bayes Classifier (NBC) and Random Forest. The results show that Deep Learning models perform better with an overall accuracy of 75\%. The results indicate that if this process is automated by applying Deep Learning for text analysis, the resultant models can exploit accident narratives which can prove to be useful for safety engineers.

\subsubsection{Text Classification Method based on Convolution Neural Network}

For better representation, and to solve the problem of manual feature extraction used in traditional methods, this paper \cite{li2017text} uses TF-IDF vectorizer to calculate and assign weight to each word in a corpus. Subsequently, it assigns weight to the word vectors as per the TF-IDF value. This process generates a text vector matrix which acts as input to the Convolution Neural Network (CNN), such that the CNN will automatically extract text features. Two distinct datasets are used for this study: NetEase News data and Text Classification data from Fudan University. The \textit{Bag-of-Words} model obtained from LSI and TD-IDF is fed into KNN and SVM. The results show that on an average TF-IDF produces better results. The classification accuracy of CNN with TF-IDF on the two datasets is 96.28\% and 96.97\% respectively. When compared to traditional classification algorithms, CNN outperforms KNN and SVM.

\subsubsection{Automated Patent Classification using Word Embedding}

This paper \cite{grawe2017automated} implements a method which uses Word Embedding and Long Short-Term Memory network to classify patents down to the subgroup. It is a process of automatic patent classification, which also compares the effectiveness of current in-use Deep Learning techniques for text processing such as Word Embedding with Word2Vec and Deep Neural Network with Recurrent Neural Network, more importantly the Long Short-Term Memory (LSTM) network. The results show that the accuracy of SVM for classifying subclass is 41\% and whereas for different levels of input, the average accuracy of LSTM is greater than 55\%. Using this result, it can be concluded that the LSTM network performs better if we increase the iterations. Additionally, instead of using a single-layer LTSM network, if we use multi-layer or grid LSTM network, we can get better results.

\subsubsection{Investigating LSTM with k-Max Pooling for Text Classification}

This paper \cite{shu2018investigating} proposes a method which uses Spatial-Dropout1D, LSTM and k-Max Pooling. The study states that word-level inputs are better than character-level inputs. Document modeling issues can be solved by using natural language comprehension which has many tasks namely: sentiment analysis, discourse analysis, paraphrase detection, machine translation, entailment recognition, summarisation, grounded language learning and image retrieval. Since the word-level approach does not have to deal with millions of distinct characters, CNN turns out to be superior even when used with only a manageable number i.e. 30000 of the most frequent words. These experiments need large datasets to train the model, hence, we used DBPendia, Yelp Review Polarity, Yahoo! Answers, Amazon Review Full and Amazon Review Polarity. The process consisted of applying baseline models and Deep Learning models, and ultimately evaluating results. The results show that LSTM k-Max pooling has better results. On the other hand SpatialDropout can give text classification regularization.

\subsubsection{Sentiment Analysis of Movie Reviews using Deep Learning and Random Forest}

This paper \cite{zharmagambetov2015sentiment} exhibits a modern approach to perform Sentiment Analysis of movie reviews by using Deep Learning Recurrent Neural Networks and Random Forest. The dataset consists of 50,000 IMDb movie reviews specially selected for sentiment analysis. Random Forest and Deep Learning using Recurrent Neural Networks is used for review classification. The \textit{Bag-of-Words} model with Random Forest gives an accuracy of 85.2\% and Deep Learning (Recurrent Neural Networks) gives an accuracy of 90.3\%. This study shows that surface analysis of text is not adequate for Sentiment Analysis. Of course, simple methods can be used for applications where high speed and resources are important. But for result-oriented applications, it would not be enough information.

\subsection{Machine Learning}

\subsubsection{Using Document Classification to Improve Performance of Plagiarism Checker}

In this study \cite{sornsoontorn2017using}, journals and thesis papers from an existing database are classified into different categories to increase the performance of Plagiarism Checker. Few pages are assigned a label and the original page which needs to be compared, is compared against a relevant paper. This is possible because a set of probability values are assigned based on the likelihood of the paper belonging to a particular category. This allows the plagiarism checker to narrow down the search space by only examining related categories and not check the entire database. The dataset consists of thesis papers and journals from Kasetsart University ranging from the year 1998 to 2010. The study used a number of established Machine Learning Classification algorithms: Logistic Regression, Naive Bayes, Perceptron, SVM, Decision Tree, Random Forest, Ada Boost, Extra Trees, and k-Nearest Neighbors. The results obtained from the top-performing models show that SVM had an accuracy of 75\%, Naive Bayes had 80\% and Random Forest had 85\%. Additionally, the study found that as the number of Decision Trees in Random Forest were increased, the model performed better but with an obvious trade-off i.e. the computation time. By increasing the trees, the accuracy of Random forest increased to 92.7\%. This lead to the conclusion that Random forest outperformed all the other classification algorithms.

\subsubsection{Identifying Expressions of Emotion in Text}

This research \cite{aman2007identifying} deals with Sentiment Analysis of text that can help determine the opinions and intent of writers, as well as their attitude and inclinations with respect to various topics. This study used blog posts as data and categorized the emotions of the posts into six different categories. For feature selection, publicly available lexical resources -General Inquirer and WordNet-Affect- were used. The algorithms used for classification were Naive Bayes and Support Vector Machines (SVM). Both, SVM and Naive Bayes, were separately modeled in combination with the feature selection methods (i.e. General Inquirer and WordNet-Affect). It was observed that by using all the features (including the non-lexical features), Naive Bayes gave an accuracy score of 72.08\% and SVM gave an accuracy score of 73.89\%.

\subsubsection{Text Classification of Short Messages (Detecting Inappropriate Comments in Online User Debates)}

This study \cite{Anthon2017text} proposes a method which involves performing text analysis when a user posts a comment on a certain article. The model analyses the comment and classifies whether it is good or bad. This study used Natural Language Processing to identity inappropriate content. Additionally, it compared a group of classification algorithms by evaluating them against different feature sets to get the best accuracy score. For this study, two datasets were used: a Swedish language dataset manually labelled by two students at LTH and an English language dataset from a Kaggle competition whose objective was to detect insults in social commentary. Multiple feature selection methods were applied to a variety of Machine Learning algorithms in different combination. It was observed that SVM and Logistic Regression demonstrated similar performance. However, Logistic Regression performed marginally better as it has the advantage of providing a probability score to show certainty that a comment belongs to one of the two classes.

\subsubsection{Mining Political Web Logs for Sentiment Classification}

This project \cite{durant2006mining} deals with the Web-Logs and discusses the prevailing techniques, and their application, for sentiment classification. The data was extracted from a political Web-Log. Naive Bayes and Support Vector Machine are the two algorithms of choice used for sentiment classification. The experiment was conducted by creating 5 collections that would act as input for the classification algorithms. The first four collections were classified by Naive Bayes and the last collection was classified by Support Vector Machine (SVM). Each collection is used to evaluate the effectiveness of a known aspect of sentiment classification. On an average, the accuracy score obtained for Naive Bayes was 78.06\% and SVM gave an accuracy score of 75.47\%. The Naive Bayes classifier performed well while SVM performed adequately.

\subsubsection{Classification of Web Documents using Naive Bayes}

This research \cite{wang2003classification} focuses on developing a system which automatically classifies a text document based on categories relevant to the document. It uses Library of Congress (LCC) to classify Web Data which is in the form of HTML. The system has three components: First is the Natural Language Processing component which parses the Web document and assigns syntactic and semantic tags. The Second component is responsible for knowledge base construction. It builds a knowledge base of information that includes the Library of Congress (LCC), subject headings, their  inter-relationships and any other information used during classification. the third and final component takes care of index generation. It generates  a set of candidate indices for each document in a test set of documents. This project implemented two different configurations of the Naive Bayes model: the Multi-Variate Bernoulli event model and the Multinomial event   model. Two different smoothing methods were also tested: Additive Smoothing and Good-Turning Smoothing. Four feature selection methods were used in the Web-Doc system:  Inverse Document Frequency (IDF), Information Gain (IG), Mutual Information (MI), and Chi-Square. A total of 722 documents (downloaded from the web) were used to perform the above experiment. When compared to previous versions of Web-Doc, we observed a 20\% increase in the F-measure (i.e., 67.19\%). The Web-Doc delivers more promising results when compared to other document classification systems. The Multinomial event model had a better performance than the Multi-Variate Bernoulli event model. Smoothing methods increased the Recall value of the classifier, but decreased the Precision.

\subsubsection{Prediction of Stock Trends using Sentiment Analysis of NEWS}

This project \cite{kalyani2016stock} implements Text Classification to classify NEWS Data and subsequently predict the trend of stocks. The dataset used is the Apple Inc.'s company data for past three years (from 1 Feb 2013 to 2 April 2016). The \textit{Bag-of-Words} model is used for text mining and sentiment prediction. Positive and Negative words were collected and compared with words of the Dataset. On comparison and analysis of the data, a word was classified to be positive or negative. SVM, Random Forest and Naive Bayes were the models used for classification. Random Forest performed very well for all test cases with accuracy values ranging from 88\% to 92\%. SVM followed closely with an accuracy score of 86\%. The Naive Bayes algorithm gave an accuracy score of 83\%. Predicting stock involves factoring in a lot of volatile features. For the above researched, it was assumed that NEWS articles and Stock Prices are related to each other.

\subsubsection{Advanced Text Analytics and Machine Learning for Document Classification}

This project \cite{anne2017advanced} handles patent document classification by classifying documents into 15 different categories and subcategories. The process revolves around developing a set of automated systems to manage and market the portfolio of intellectual properties while facilitating easier discovery of relevant intellectual properties. The Data was retrieved from NASA patent documents which are in HTML format and can be downloaded from U.S patent office. k-Nearest Neighbors (kNN), two variations of the Support Vector Machine (SVM), and two tree-based classification algorithms: Random Forest and J48 are the modela used for the classification. The following accuracy scores were achieved: KNN: 54\%; SVM (RBF kernel): 55.4\%; J48: 57.07\%; Random Forest: 61.04\% and SVM (Poly kernel): 69.2\% (96.03 after removing unclassified data). The study includes an interesting finding that that removal of unclassified data increases the accuracy from 69.2\% to 96.0\%.

\subsubsection{Supervised Text Classification of Medical Triage Reports}

This thesis \cite{kleverwal2015supervised} focuses on categorizing Triage complaints based on symptoms entered in the interface. The process is divided in two parts. In the first part, the limitations of the current Triage system are described. In the second part, an approach based on Supervised Text Classification is implemented and evaluated . The data consists of unstructured texts and related complaints entered by the officer in the Triage interface. The Regionale Ambulance Voorziening Utrecht(RAVU), an organization that coordinates all emergency care and ambulance transportation in the Utrecht province of the Netherlands made this dataset available for research. The query construction methods, OR-query and Unigrams are compared to different Nearest Neighbor’s values. The query construction methods, Nearest Neighbors and scoring methods are common modifications of the kNN algorithm. There is no significant improvement in performance as compared to current practices. Therefore, there is no need for replacement of the current system. 

\subsubsection{Text Classification-Classifying Events to Ugenda Calendar Genres}

This thesis \cite{crijns2016text} proposes a model to automate the process of information management-validating and preparing data for venues- in the Ugenda website, as it is a very time-consuming process. Automatic Text  Classification has a long way to go as it has not been optimized enough to  challenge a human’s capabilities \cite{kaushik2016comprehensive}. Text classification is usually divided into two broad tasks: the first task is pre-processing; and the second task is classification which is used to predict a label for unseen events \cite{crijns2016text}. Support Vector Machine, Logistic Regression, Naive Bayes and Random Forest are the models used for classification. The Logistic Regression classifier delivers high performance with an F1 Score: 0.8110. Further comparison lead to the result that SVM and Random Forest show similar results, and Naive Bayes is less accurate. The genre which has the largest support has the best performance. Other labels with a much smaller support also perform well. Genres that contain ambiguous events and miscellaneous genres do not perform well \cite{crijns2016text}.

\subsubsection{Automatic Web Page Categorization using Text Classification Methods}

This study used Natural Language Processing and Machine Learning to implement Text Classification for Web-Page Categorization \cite{tobias2017automatic}. The process involved extracting textual, natural language content from web-pages and encoding the document as a feature vector with natural language processing methods. The data for this study was extracted from ‘Whaam’, a platform on which a user stores links of different categories. Data in a web-page is divided accordingly as per development of the web-page and classification models are applied on every division. Support Vector Machines (SVM), k-nearest neighbor (kNN), Multinomial Naive Bayes (MNB), Term-Weighted Complementary Naive Bayes (TWNCB) and the Cavnar-Trenkle N-Gram-based classifier (N-Gram) were the algorithms used for classification. The results show that they are not terrible, suggesting that honing of these models will increase the success rate.

\subsection{Deep learning Models}

\subsubsection{Commented Content Classification with Deep Neural Network based on Attention Mechanism}

This paper \cite{zhao2017commented} proposes a CNN-Attention network based on Convolutional Neural Network with Attention (CNNA) mechanism in which all information between words, for context, can be expressed by using different sizes of convolution kernels. Additionally, an attention layer is added to the CNN to obtain semantic codes which has the attention probability distribution of input text sequences. Next, weights of text representing information are calculated and finally, SoftMax is used to classify emotional sentences. This model was proposed because it is very difficult to represent text information with shallow network, and it is time-consuming to use deep neural network. In this method the features of different context information can be extracted thereby reducing the depth of the network and hence, improving the accuracy. The dataset used is the COAE2014 task 4 weibo dataset. The results show that Text semantic information and Rich Text features are extracted using bottom layer and experimental results show that the accuracy of the proposed model is 95.15\%.

\subsubsection{Classification of Short-Texts Generated During Disasters: A Deep Neural Network based Approach}

This paper discusses that recent advancements in social networking can be used to provide help to the people in need \cite{kundu2018classification}. In case if anyone needs blood, gadgets etc., the go-to approach is going on social network. This paper proposes a Deep Learning based model to classify the tweets into different actionable classes on the basis of resource need, availability, activities of various NGO etc. at the time of an ongoing natural disaster. During disaster events, these micro blogging site provide a rich source of data in the form of short texts. Classification of data into different classes helps in differentiating people who are in need and/or can be rescued. This paper states that previous works don’t factor in semantic information available in the text. Hence, this study utilized word embeddings to incorporate semantic information, before giving input to the model. The dataset used is obtained from Forum for Information Retrieval Evaluation 2016 (FIRE2016) and 2017 (FIRE2017). The FIRE2016 dataset has approximately 2100 labelled tweets belonging to 7 classes and the FIRE2017 dataset has approximately 900 tweets belonging to 2 classes. The paper compared the LSTM model with the 5 different models and the results show that the performance of LSTM when applied on the FIRE2017 dataset is superior than all other models in terms of Precision, Recall and F-Score. For the FIRE2016 dataset, CNN performs better than all other models in terms of Precision, but at the cost of Recall which brings down its F-score. On the other hand, the proposed approach performs moderately for Precision, but produces a significantly better value for Recall. The paper concludes that the proposed method has best value of the final F-Score out of all the methods which were compared. The results obtained from the FIRE2017 dataset are significantly better when compared to that of the FIRE2016 dataset. This is due to the fact that FIRE2017 contains only 2 classes whereas FIRE2016 has 7 classes, making it a more complex classification task. Additionally, It can be concluded that the use of word embeddings as input, and the automated learning of useful features during the training process makes the Deep LSTM model perform the best.

\subsubsection{A Deep Learning based Ultrasound Text Classifier for Predicting Benign and Malignant Thyroid Nodules}

This project deals with the use of Deep Learning methods in computer-aid diagnosis, by predicting Benign and Malignant Thyroid Nodules \cite{chen2017deep}. Traditional pre-operative diagnosis has two stages: an Ultrasound check and a Fine Needle Aspiration. This traditional approach of Fine Needle Aspiration is in trouble when in practice since, there is a potential risk of turning Benign Nodules into Malignant ones. The Malignant Nodules are further confirmed by surgery and pathology. This paper proposed a Deep Learning based Ultrasound Text Classifier for predicting Benign and Malignant Thyroid Nodules. A structured analysis has been done based on the Dependency Syntax Analysis. In this project, the dataset used is real medical data. Additionally, the new Thyroid dataset downloaded from UCI was compared.  The proposed method, based on Deep Neural Network, was applied on labelled Ultrasonic Text with Benign or Malignant label of pathology. The results of the proposed methods were compared to the base models. The results show that the proposed method produced the highest accuracy rate of 93\% and 95\% for both the real medical dataset as well as the UCI standard dataset. Further, traditional classification models: Random Forest, Support Vector Machine and Neural Network were applied on the real medical dataset as well as the UCI standard dataset. The proposed Deep Neural Network outperformed all the other models and the result is convincing even when we take the real medical data imbalance into account.

\subsubsection{Aksara Jawa Text Detection in Scene Images using Convolutional Neural Network}

This thesis \cite{afakh2017aksara} deals with the preservation of Aksara Jawa text in the loss heritage culture. Aksara Jawa text is an ancient Javanese character mostly written on stones to describe history or naming, such as places, wedding, tombstones, etc. This study used Deep Learning models such as Convolutional Neural Networks to localize the existence of Aksara Jawa text. A text detection system in Scene Images was developed by embedding this CNN method. The existing method used manually designed features, and moreover, the classifier is explicitly applied on it. The proposed technique allows the classifier and features to jointly learn and subsequently, back-propagation is employed to simultaneously obtain parameters. The dataset used is the CDAR 2003 training set. The synthetic data is mixed with Aksara Jawa font text and scene images collected from the internet. CNN produced a precision values of 0.96. This was significantly greater than previous works which have a precision value of 0.63. Additionally, it was observed that Aksara Jawa text can be classified in extremely complex images which have dark or blurred backgrounds. The experiment was efficient and exhibited encouraging results. For future work, the paper proposed that the Neural Network could be intertwined with more layers to produce a Deeper Network which could potentially result into a higher accuracy score.

\subsubsection{Text Detection in Natural Scenes using Fully Convolutional Dense Nets}

This study states that over the past couple of years Convolutional Dense Nets have been extremely successful in Object Detection and Recognition, yet they have never been used to detect Text. To understand the semantics in a natural scene, classification of these texts by detection and recognition are important steps. In this study, a semantic segmentation was trained on images which segmented each image into 3 sections: text, background and word-fence \cite{behzadi2018text}. Word-fence was used to separate close words in the model that were being learned. The proposed method also used synthetic data to train and real data for tuning. The dataset used for this study contains 200,000 instances of SynthText; ICDAR Multi-Lingual-Text dataset with 7200 images to fine-tune the model; and ICDAR 2013. The proposed method used Fully Convolutional DenseNets which was based on semantic segmentation of the images. They results showed that the proposed method achieved acceptable results and it can be a viable approach for text detection.

\subsubsection{Spam detection in Online Social Networks using Deep Learning}

The aim of this thesis \cite{ameen2018spam} was to detect Spam in social networking sites by using Deep Learning models. In this study, Word2Vec was used for feature selection and subsequently, Deep Learning models were used to distinguish spam from the non-spam tweets. The tweets, for conducting this study, were collected over 5-days using Twitter’s Streaming API. The dataset contained approximately 10 million tweets. The tweets were processed using JSON designing and every line in the structure was handled as a target. Different measures such as Precision, Recall, and F-measure were used to evaluate the performance of the methods. Multilayer Perceptron, Random Forest and J48 were the algorithms used for the classification. The results showed that the precision value of the Multilayer Perceptron (MLP) was 92\%. The worst result for this measure was obtained from the Naive Bayes classifier. The study on these tweets exhibited that Deep Learning models outperformed conventional and classical classification models.

\subsection{Hybrid Models}

\subsubsection{Electronic Medical Data Analysis based on Word Vector and Deep Learning}

This paper studies and implements combination of Word Vector and Deep Learning to handle the issue of repetition, missing semantics etc. in Electronic Medical data \cite{li2018electronic}. In this paper, an integrated learning method is implemented. The process involves training classical learners on the original dataset, the output is used as the new dataset to train again to get Meta-Learner. Models such as LSTM-improved model and Yoon-improved model are trained, and their output is used as stated earlier. From the experimental results it was found that among the traditional methods, SVM had the highest accuracy score of 83\%. Out of Lenet Network, LSTM-improved model and yoon-improved model, the LSTM-improved model outperformed other models with an accuracy score of 93.4\%. As the paper proposed an integrated learning approach, the implementation was extended and the LSTM-improved model and the Yoon-improved model were integrated which gave the highest accuracy of 94.7\%. The pros and cons of each individual technique was accounted for, there by retaining a greater degree of semantic information and as a result achieving a better accuracy value.

\subsubsection{Sentiment Classification: Feature Selection based Approaches versus Deep Learning}

This paper compared the performance of Feature Selection based techniques and Deep Learning (Convolutional Neural Network, Long Short-Term Memory Network, and Long-Term Recurrent Convolutional Network) \cite{uysal2017sentiment}. For text classification, traditional methods of Feature Selection such as \textit{Bag-of-Words} was used. The newly transformed features from \textit{Bag-of-Words} was given as input to classical models such as Support Vector Machines (SVM) and Neural Networks. The disadvantage of \textit{Bag-of-Words} is that there is a significant loss of data during the processing stage. Since the model is based around the frequency of a word, it doesn’t take into account the importance and order of the words with low frequency. In order to overcome this problem a novel technique called Word Embedding was developed for Text Mining. This paper conducted an experiment on 4 datasets with different characteristics: IMDB movie review dataset; subset of Sentiment140 dataset; Nine Public Sentiment dataset consisting of positive and negative reviews from domains such as laptop, music etc; and Multi-domain dataset which contains reviews collected for different products from Amazon.com. For analyzing Deep Learning techniques, Convolutional Neural Network, Long Short-Term Memory Network, and Long-Term Recurrent Convolutional Network were implemented. For the Feature Selection approach, a Hybrid method, which combined \textit{Bag-of-Words} and Averaged Word Embedding, was used. Firstly, SVM was used to classify both the feature selection models. Subsequently, Deep Learning models and the proposed Hybrid Deep Learning models are used. The results showed that at least one Deep Learning Model outperformed all Feature Selection based methods. The Hybrid model, which was a combination of CNN + LSTM, was the best performing mode for the IMDB dataset. CNN performed the best for the Nine Public Sentiment and Multi-Domain dataset.  The Sentiment140 was the only dataset on which IG and Word Embeddings gave better results. It was concluded that better results can be achieved in Sentiment Classification by using Deep Learning models with fine-tuned Semantic Word Embeddings.

\subsubsection{A Hybrid Deep Learning Model to Predict Business Closure from Reviews and User Attributes using Sentiment Aligned Topic Model}

In this study, a Hybrid Deep Learning model is developed to predict the potential close of a business with in a specific period \cite{thazhackal2018hybrid}. Business closure is an indicator for success or failure of a business which helps investors and banks to make decision whether to invest in a venture for future growth and benefits. This study used a bookstore dataset obtained from Yelp. Classical methods such as Decision Trees, SVM, Random Forest and Gradient Boosting, with optimal configuration, were applied. Additionally, Deep Learning models such as Deep Neural Networks and Convolutional Neural Networks, along with the Transfer Learning was applied. This paper experimented by using different combinations of CNN, DNN and Transfer Learning. The proposed method of this paper was a Hybrid Method which used both Data Frame and Aspect-Wise ratings as input. This method was also implemented in conjunction with Transfer Learning. The results obtained from the classical model on the dataset were not promising. On the other hand, the results achieved from the proposed model were significantly better and outperformed the previous work. Additionally, it was observed that incorporation of Transfer Learning increased the accuracy in all Hybrid combinations. It was concluded that word of mouth in a business plays a crucial role for deciding the sentiment score and the parameters which ere used should play a major role in improving the performance.

\subsubsection{Target Advertising Classification using Combination of Deep Learning and Text Model}

Now-a-days, there is a boom in Online Advertising. Online Advertising is not only preferred for increasing the sales of a product and promote services, but also to build a successful brand image. As a consequence, filtering of content became necessary in order to detect improper information before it gets advertised on websites and social media. The dataset used in this study \cite{phaisangittisagul2019target}, was collected from users who posted Promotional advertisements at www.wongnai.com. The first phase of the project dealt with Image Classification. The image was classified on the basis of whether it was related to the advertisement or not. The second phase of project was Text Classification, which is a common task in Natural Language Processing (NLP). The promotional advertisement was classified as a Good or Bad advertisement by checking the number of bad words. Finally, the results of the Image and Text classification were integrated as a final prediction from the model. The results showed that manual classification by humans gave an accuracy score of 82.95\%, whereas the proposed methodology gave an accuracy of 90.15\%. It was concluded that based on the combination of Image and Text classification, the performance achieved using the proposed methodology was more satisfactory as compared to manual (human) classification. However, there is room for improvement for the proposed model.

\subsubsection{A Neural Network-based Ensemble Approach for Spam Detection in Twitter}

This thesis focused on a Neural Network-based Ensemble approach for detecting Spam in Twitter \cite{madisetty2018neural}. In this study, 5 CNNs and one feature-based model were used in the ensemble. For the CNN, different Word Embeddings (Glove, Word2Vec) methods were used to train the model. The Feature-based model on the other hand, used content-based, user-based, and n-gram features. The proposed model was the combination of both the Deep Learning models and the traditional Feature-based models. The model used a Multi-layer Neural Network which acted as a Meta-Classifier. Two datasets, one balanced and one imbalanced, were used to evaluate the performance of the model. The results showed that a CNN with any kind of embedding gave an accuracy score greater than 90\%. Random Forest achieved an accuracy of approximately 82\%. The proposed ensemble model achieved an accuracy score of approximately 95\%. After performing all text on all datasets with different combinations, it was concluded that the performance of the proposed ensemble approach was found to be superior than the baseline methods.

\subsubsection{Research on Hot News Classification Algorithm based on Deep Learning}

Now-a-days people are more interested towards Hot news because of advancements in the arena of Network Applications. Hence, there is a need to classify the said Hot News. This study investigated various Deep Learning approaches in Text Categorization, combined with characteristics of news text, and put forward a double Bi-Gated Recurrent Unit (GRU) and Attention Deep Learning model to predict Hot news \cite{wang2019research}. Classical Machine Learning classifiers such as Naive Bayes, Support Vector Machine (SVM), Classification tree and Vector-based features cannot study the semantics and hence, the classification accuracy is not high. The above issue and associated drawbacks is the reason why sub-field classification has become more challenging. This paper proposed a novel Hybrid Deep Learning model, namely BiGRU and Attention Mechanism, which can be used for text categorization. The data was fetched from the sohu website crawl (from October 20, 2015 to April 25, 2016). It included financial news, social news and other Hot news from the sohu buzz. The experimental results showed that the predicted results of the proposed model were more promising than the traditional Deep Learning model. The proposed model achieved a higher Precision when compared to classical text classification models. This was mainly due to the problem of focus and inability of classical models, to consider context in text data.

\subsubsection{A Hybrid Deep Learning Model for Text Classification}

The objective of this study was to develop a hybrid model which combined Bidirectional Gated Recurrent Unit with attention (BiGRU), Convolutional Neural Networks (CNN) and Attention mechanism \cite{chen2018hybrid}. Since the classical feature selection methods such as the \textit{Bag-of-Words} model and \textit{Vector Space} model leads to significant loss of information, Word Embedding was introduced in order to overcome this issue. It is based on Deep Learning models and is widely used by researchers where they use it in conjunction with Convolutional Neural Networks (CNN), Recurrent Neural Networks (RNN) and Attention mechanism. The data for this study was collected from the NLP\&CC2018 task. The results obtained from the Hybrid model were compared to other models such as CNN, RNN, RCNN and GRA. From the results, it was observed that the proposed Hybrid model achieved more promising values for Precision, Recall and F1-score when compared to the other methods. This proved that the proposed Hybrid model was more effective in Text Classification. Additionally, it was observed that the results of RCNN and GRA are better than the traditional CNN and GRU, which indicated that a Hybrid model can overcome the learning ability of a single Deep Learning model to a certain extent.

\subsection{BERT (Bidirectional Encoder Representations from Transformers)}

The novel NLP model BERT was conceived in late 2018 by researchers at Google AI Language. It has disrupted the Machine Learning community by presenting promising results in virtually every NLP task. BERT's state-of-the-art feature is applying the bidirectional training of Transformer to language modelling. BERT utilizes Transformer, an attention model that learns contextual relations between words in a text. Transformer is made up of two components:
\begin{itemize}
    \item an Encoder: Reads text input
    \item a Decoder: Produces a prediction for the task
\end{itemize}
For BERT, only the Encoder is necessary since the goal is to create a language model. The Transformer encoder accepts a sequence of tokens as input. Firstly, the tokens are embedded into vectors and subsequently these tokens processed in a Neural Network. The output from the Neural Network is a sequence of vectors where each vector corresponds to the input token at that index. BERT performs classification tasks (e.g. Sentiment Analysis) by adding a classification layer on top of the Transformer output layer for the Class tokens. Results obtained from BERT have shown that a bidirectional trained language model has a deeper sense of language (context and flow) than unidirectional language model.  

Ever since the inception of this breakthrough NLP model, researchers have invested time and resources in exploring this novel technology. A method for Automatic text classification based on BERT was proposed in a study conducted by the Huaiyin Institute of Technology, China. \cite{8975793}. In the study, BERT is used to transform text-to-dynamic character-level embedding, and BiLSTM (Bi-Directional Long-Short Term Memory) and CNN output features are combined to better represent the text and consequently, improve the accuracy of text classification. The proposed method combining BERT, BiLSTM and CNN resulted into an Accuracy score of 96\% for text classification of 10 labels.

Another notable work using BERT was demonstrated by researchers at the National Institute of Technology, Tokyo College, Japan \cite{8959920}. This study developed an Automatic abstractive text summarization algorithm in Japanese using BERT and LTSM (Long short-term memory). In this study, BERT is used as the Encoder to produce a feature-based input vector of sentences which is passed on to the Decoder (i.e. LTSM)  which returns the summary sentence. Results of this paper revealed that the model was successful in learning as it captured key points of the text in the summary sentence. However, it was observed that the proposed model was unable to handle unknown words which lead to repetition of words in certain sentences. 

\section{Methodology}

In order to provide credibility to the observed results and to make the research reproducible, this section presents the general information about the dataset, the experimental methodology and settings used for each classification method.

\subsection{Datasets}

\begin{figure}[htb]
\centering \includegraphics[width=7.5cm, height=6cm]{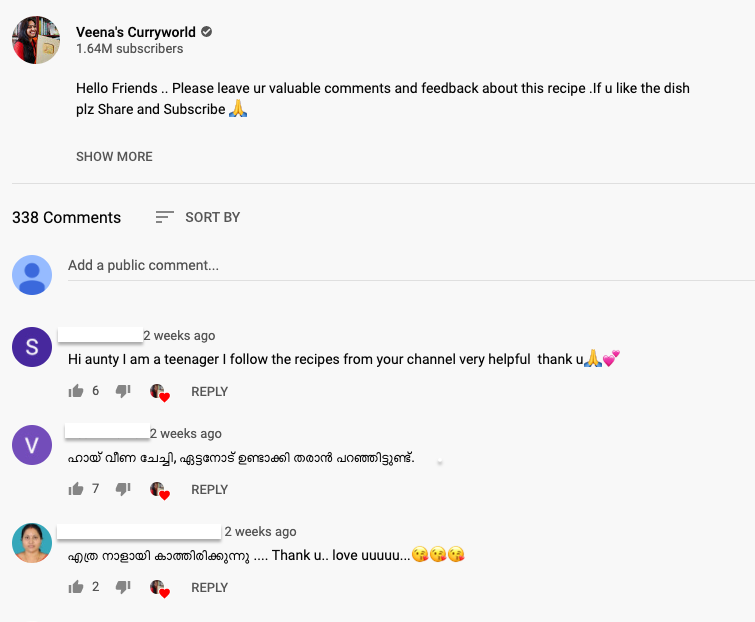}
\caption{\textbf{Veena's Curryworld}}
\label{figure1}
\end{figure}

\begin{figure}[htb]
\centering \includegraphics[width=11.2cm, height=6cm]{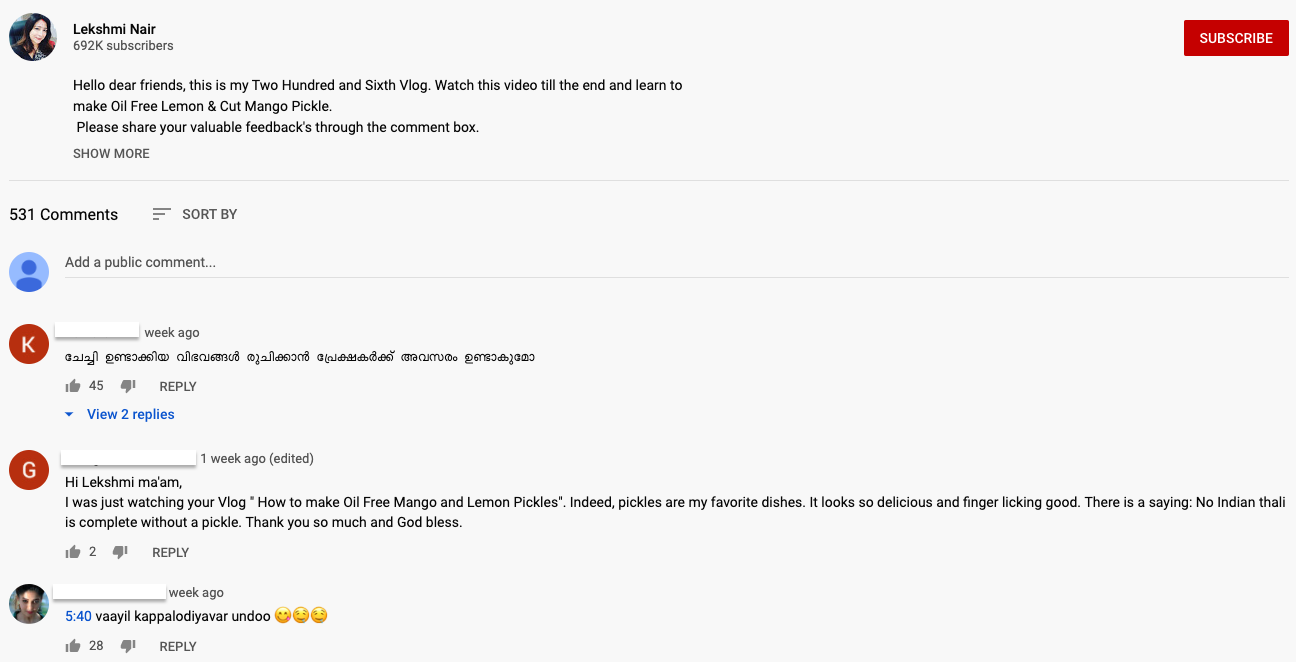}
\caption{\textbf{Lekshmi Nair}}
\label{figure2}
\end{figure}

The dataset \cite{kazhuparambil_subramaniam_2020_3871306} used for performing text classification is a combination of two distinct datasets scrapped from the comment section of two YouTube cooking channels namely \textit{Veena’s CurryWorld} \footnote{https://www.youtube.com/channel/UCugwC12zllpi9zJoUP5ZVyw} and \textit{Lekshmi Nair} \footnote{https://www.youtube.com/channel/UC2zTRxJjjNYZpdXHk\_blEHg}. It consists of 4291 comments. The data is real, public and non-encoded, and extracted directly using the YouTube API. The dataset contains two attributes, \textit{text} and \textit{label}. Figure \ref{figure1} and Figure \ref{figure2} show sample comment snippets taken directly from the respective YouTube Channels.

\begin{table}[h!]
\caption{\textbf{Types of Comments}}
\label{table2}
\centering \includegraphics[width=14cm, height=16cm]{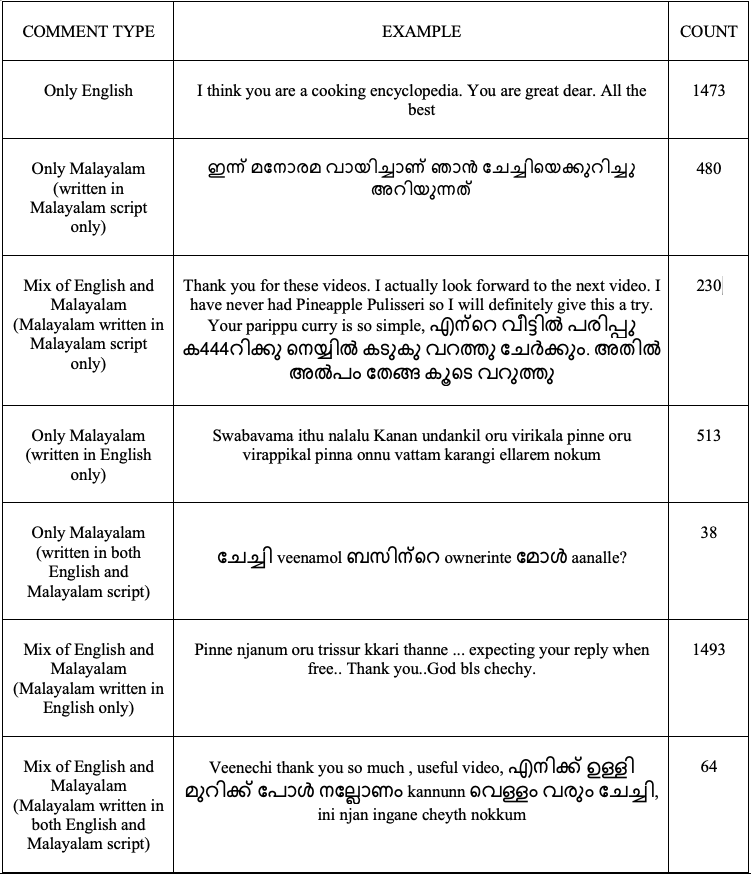}
\end{table} 
\begin{table}[h!]
\caption{\textbf{Label Description \& Frequency}}
\label{table3}
\centering \includegraphics[width=7cm, height=3.5cm]{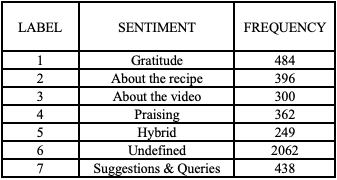}
\end{table} 
\begin{table}[h!]
\caption{\textbf{Translated Comments with Corresponding Labels}}
\label{table4}
\centering \includegraphics[width=7.8cm, height=18.5cm]{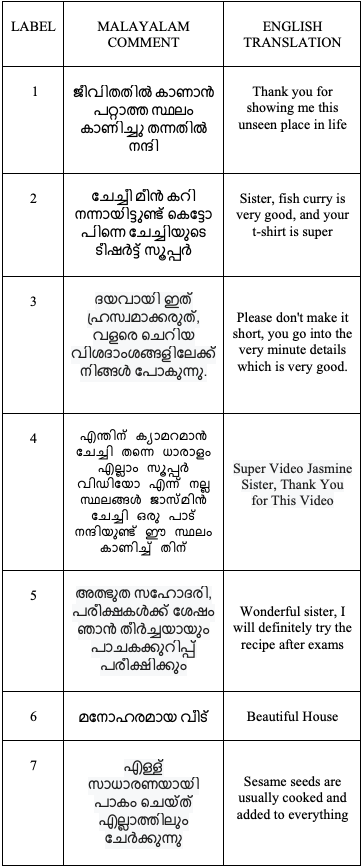}
\end{table}  
Each record represents a text comment, which is a mix of English and Malayalam, posted in the comment section of videos of the two channels. Table \ref{table2} presents the different comment types along with examples and count.

All records were manually labelled as per the expressed sentiment. Metadata information such as author’s name, date of publishing etc. have been omitted from the dataset. Table \ref{table3} presents the labels assigned to the text comment, their sentiment and corresponding frequencies observed in the dataset \cite{kazhuparambil_subramaniam_2020_3871306}. The labels were inspired from Kaur \textit{et.} al. \cite{kaur2019cooking} and Shah \textit{et.} al \cite{shah2020opinion} which also focused on comment classification. Additionally, Table \ref{table4} presents examples of comments corresponding to each label along with the English translation.

\subsection{Experimental Methodology}

The experiment is divided into two distinct phases. The first phase is based around text classification using the BERT language model. The second phase involves employing traditional machine learning classification models. The result of these phases are later compared in Section 4.

\subsubsection{Language Model Workflow}

BERT is a language model that learns to predict the probability of a sequence of words (tokens). It is an N-gram language model which computes the probability of encountering a word \textit{N} based on the joint probability of encountering previously observed \textit{N-1} words together. While language models are the industry standard for text completion, their ability to interpret flow and context, and to correctly predict and complete sentences, can be extended to supervised learning in the form of text-classification. BERT is accessible via the \textit{simpletransformers}\footnote{https://pypi.org/project/simpletransformers/} library which is built on top of the \textit{transformers}\footnote{https://github.com/huggingface/transformers} library by \textit{HuggingFace}\footnote{https://github.com/huggingface}. With the help of the \textit{simpletransformers} library, the experiment was simplified. Once the data was cleaned and randomly  sampled, BERT elegantly performed the required preprocessing tasks such as splitting, stopword removal and stemming along with feature engineering. The steps of pre-processing and feature engineering were embedded inside the training phase of the BERT language model.
BERT is a comparatively newer language model conceived in 2018 and, DISTILBERT and XLM are extensions of the same. The transformers library currently consists of 3 multilingual models which partially support Malayalam:
\textbf{bert: bert-base-multilingual-cased}; 
\textbf{distilbert: distilber-base-multilingual-cased};
and \textbf{xlm: xlm-mlm-100-1280}.
	
\begin{figure}[h!]
\centering \includegraphics[width=13.175cm, height=5.645cm]{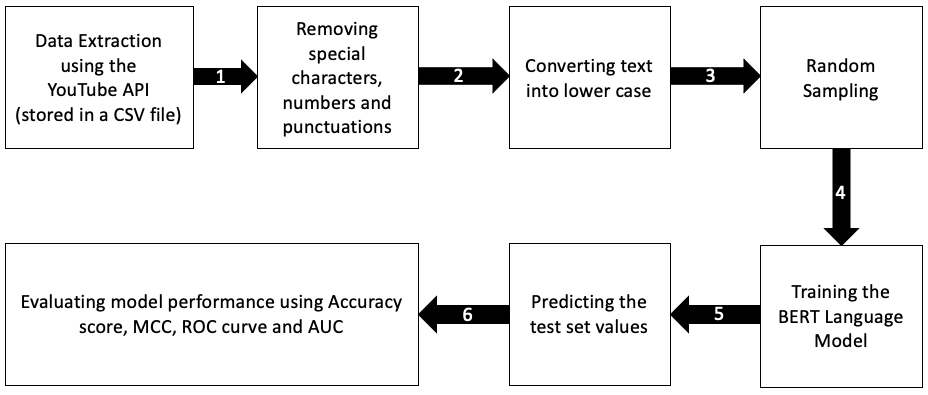}
\caption{\textbf{Language Model Workflow}}
\label{figure3}
\end{figure}

To further investigate model performance, hyper-parameter tuning was conducted. The train\_batch\_size, eval\_batch\_size, num\_train\_epochs and learning\_rate were altered to find optimal combination of values that gave the best accuracy. The optimal configuration was later evaluated on the test set using the following metrics: Accuracy score, Matthews Correlation Coefficient (MCC) and Receiver Operating Characteristic (ROC) curve and Area Under Curve (AUC). Figure \ref{figure3} shows the workflow of this phase of the experiment. 

While Accuracy score is a very intuitive metric to evaluate model performance, it is sensitive to imbalanced data especially in multi-text classification. The ROC curve provides an alternate perspective by elegantly evaluating the performance of the model with respect to its capacity to accurately distinguish between classes. This is especially desirable in multi-text classification where one wants to gain insight into how accurately the model is predicting each class. Further, the AUC gives the rate of successful classification for each class by the model. Additionally, MCC is one of the most robust metrics for multi-text classification where all the classes are of interest. MCC being perfectly symmetric, handles asymmetric and imbalanced data well. It takes into account all four values of the confusion matrix and doesn't give one class more importance than the other. Its ability to indicate positive/negative correlation when it correctly classifies/misclassifies makes it a reliable metric for multi-text classification. 
\subsubsection{Machine Learning Workflow}
The second phase of the experiment involved employing established top-performing classification models. Once the CSV file (containing all the comments and corresponding Labels) was imported and stored in a pandas data frame, the first preprocessing step was to clean the data and extract only the words from the comments. Special characters, numbers, punctuations and any form of redundant data was removed. Subsequently, the words were transformed into lower case and the comment string was split into individual words. Once individual lower case words were extracted, Stemming was conducted and stopwords were filtered out. These pre-processing steps were applied to each comment of the dataset. The processed dataset was feature engineered to convert the dataset into a \textit{Bag-of-Words} model and frequency representation, as described below.
Consider a comment c consisting a set of terms (tokens) \[c = w_1, …, w_n\] where in each term \textit{w\textsubscript{k}} corresponds to a single word with two or more alphanumeric characters. \textit{Bag-of-Words} model makes it possible to represent each message as a vector \[\overrightarrow{v} = \{v_1, …, v_n\}\] where \textit{v\textsubscript{1}, ..., v\textsubscript{n}} are attribute values \textit{V\textsubscript{1}, ..., V\textsubscript{n}} related to the terms \textit{w\textsubscript{1}, ..., w\textsubscript{n}}. The attributes are integer values obtained from the vectorizer employed, which represents how often each word occurs in a comment. 

\begin{figure}[h!]
\centering \includegraphics[width=\textwidth, height=5.645cm]{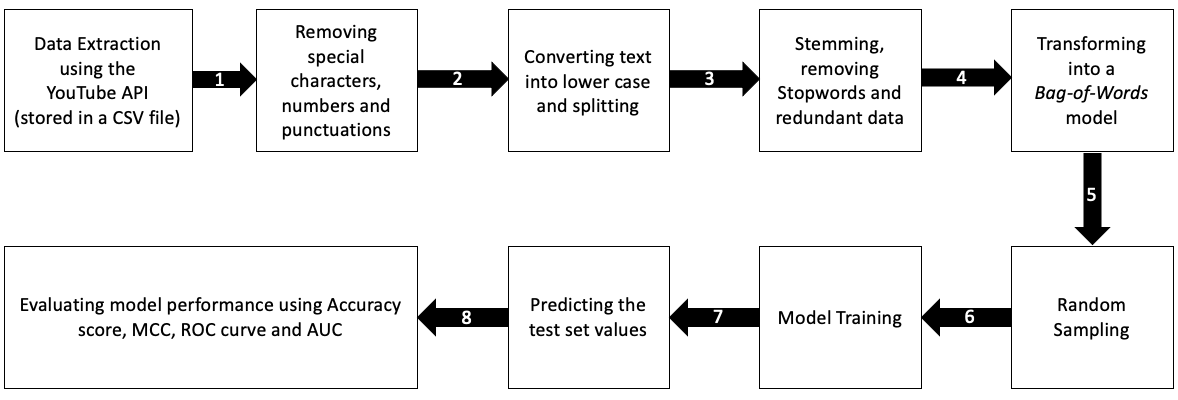}
\caption{\textbf{Machine Learning Workflow}}
\label{figure4}
\end{figure}

The following vectorizers were used for Feature Engineering: Count Vectorizer, TF-IDF (Term Frequency - Inverse Document Frequency) Vectorizer, Term Frequency Vectorizer and Hashing Vectorizer. The feature engineered data was randomly sampled into a 80/20 train and test split and the following classification methods were employed: Multinomial Naïve Bayes, K-Nearest Neighbors, Decision Tree, Random Forest and Support Vector Machine. Once the classification models were trained, they were applied on the test set to predict class Labels. The model performance was evaluated using the same metrics which were used in Phase 1 of the experiment i.e. Accuracy Score, Matthews Correlation Coefficient (MCC) and Receiver Operating Characteristic (ROC) curve and Area Under Curve (AUC). Figure \ref{figure4} shows the workflow of this phase of the experiment. 

\indent The experiment (Classification using \textit{Bag-of-Words} and BERT), in its entirety, was implemented in \texttt{Python 3.7.4} using the following libraries:
\texttt{scikit-learn (v.0.21.3)}\footnote{https://scikit-learn.org/stable/};
\texttt{transformers (v.2.5.0)};
\texttt{tensorboardx (v.2.0)}\footnote{https://pypi.org/project/tensorboardX/};
\texttt{simpletransformers v.0.20.3}. 

The \textit{k}-NN model was tested for different values of \textit{k}, where it was observed that the model performed best for \textit{k = 5}. Additionally, Random Forest was tested with values of \textit{n\_estimators = 100, 1000 and 2000}. The SVM model was trained using both the "linear" and "rbf" kernel. The results of the better performing configuration (i.e. the "rbf" kernel) were used. The remaining models models were trained using their optimal configurations. For methods with random initialization, such as Train/Test split, Decision Tress and Random Forest, the random number generator seed was set to 42 for the purpose of reproducibility.

\section{Results}

This section draws a comparison between traditional classification models and the novel transformer based language model. Performance of all the models is evaluated using the aforementioned metrics. 
Table \ref{table5} presents the results achieved by the advanced language models. The results are sorted by accuracy values. XLM was the top-performing model out of all the three language models, with an Accuracy of \textbf{67.31\%} and MCC = \textbf{0.531} at its observed optimum configuration. It was closely followed by DISTILBERT (Accuracy = \textbf{66.92\%}, MCC = \textbf{0.520}). BERT was the poorest performing model with an Accuracy of \textbf{65.82\%} and MCC = \textbf{0.519}. It was observed that increasing the \textit{learning rate} always resulted into a drastic decrease in model Accuracy. Additionally, it was noticed that it also threw an exception while computing the value of MCC. The optimum value for \textit{train\_batch\_size} and \textit{eval\_batch\_size} was observed to be 8. Further, it was noticed that increasing the two values always decreased model Accuracy. This observation aligns with the conventional practice of choosing small batch sizes. Smaller batch sizes are often easier to fit and train in memory. Additionally, small batches are often noisy which helps in reducing generalization error. This is especially desirable for the purpose of reproducibility and in order to maintain the consistency and performance of the model on unseen data. 

\begin{table}[htb]
\caption{\textbf{Results Achieved by Language Models}}
\label{table5}
\centering \includegraphics[width=9cm, height=10cm]{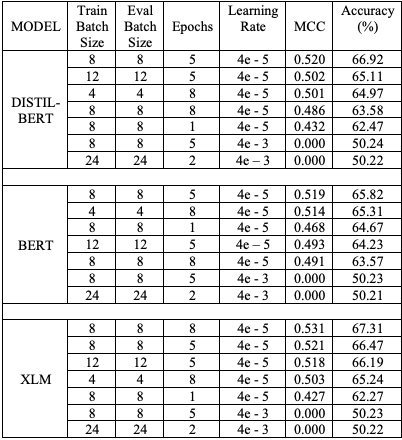}
\end{table}

\begin{figure}[!h]
\centering \includegraphics[width=\textwidth, height=10cm]{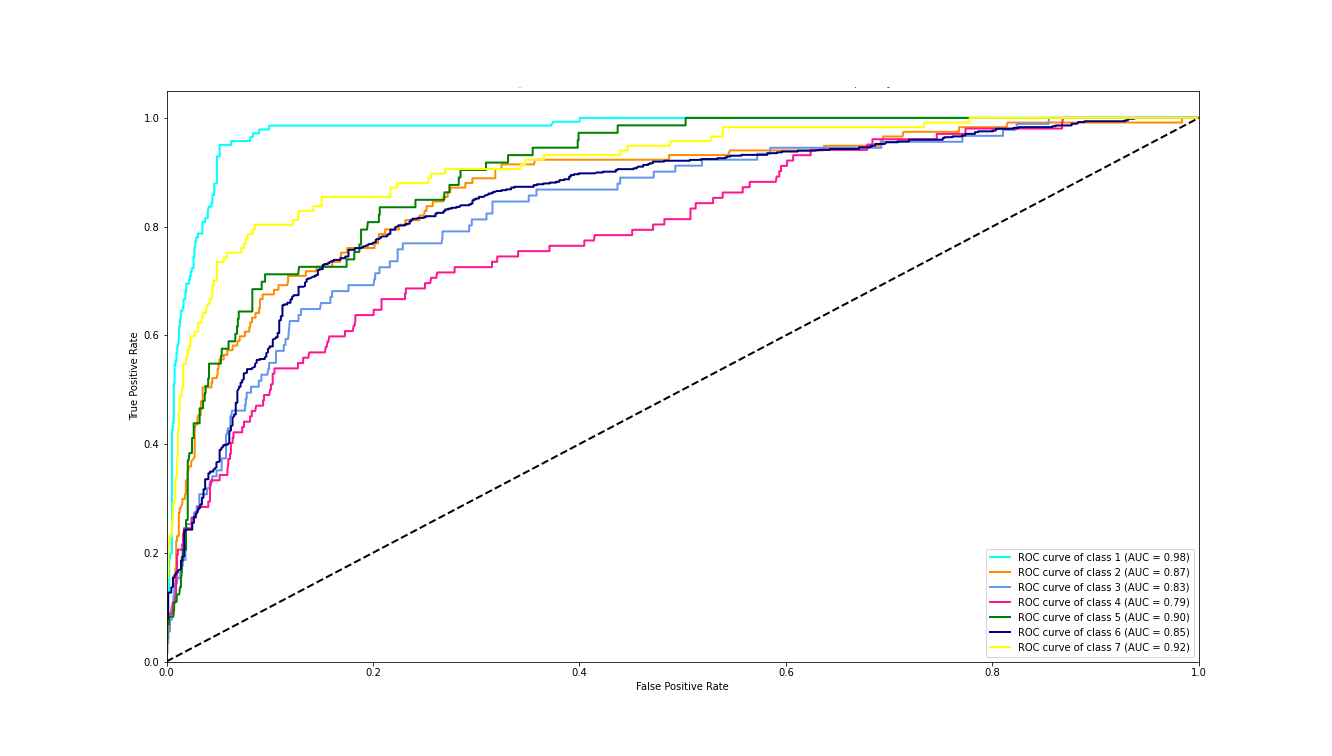}
\caption{\textbf{ROC Curve \& AUC for XLM}}
\label{figure5}
\end{figure}

\begin{figure}[!h]
\centering \includegraphics[width=\textwidth, height=10cm]{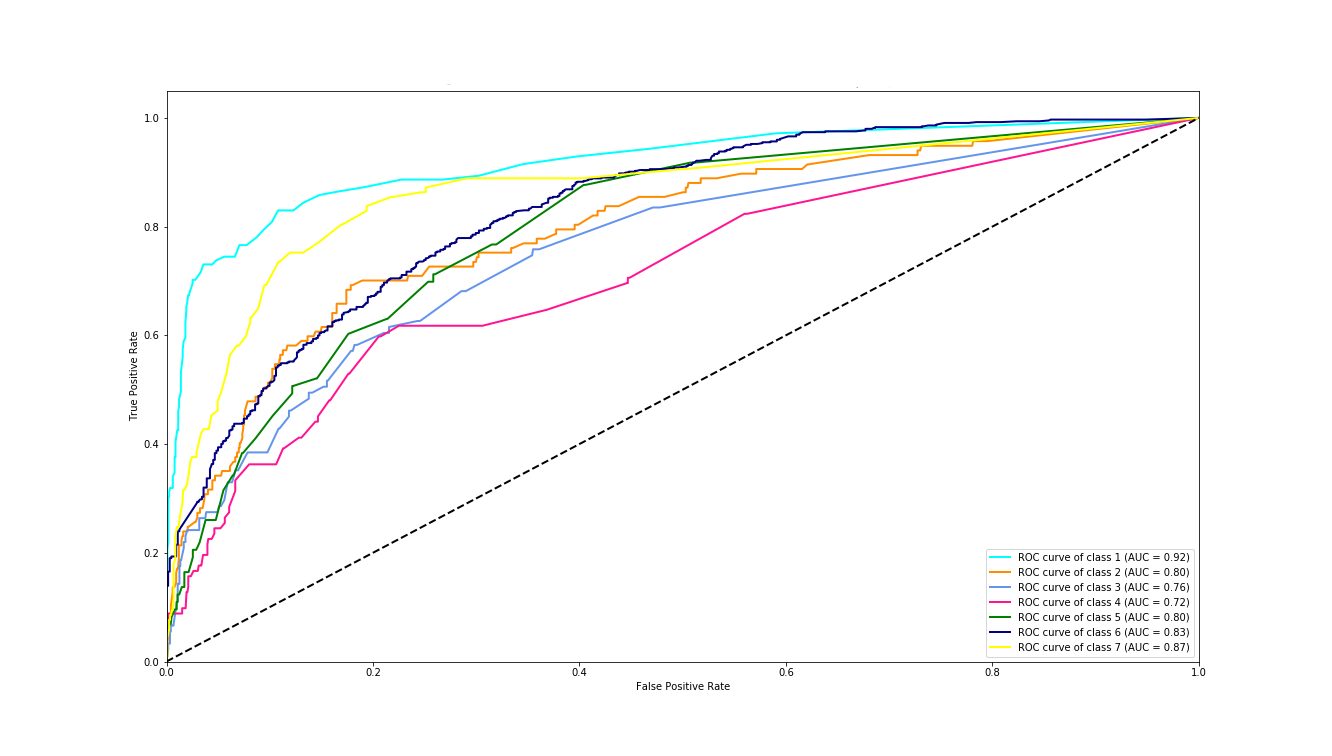}
\caption{\textbf{ROC Curve \& AUC for Random-Forest with Term-Frequency Vectorizer}}
\label{figure6}
\end{figure}

\begin{table}[htb]
\caption{\textbf{Results Achieved by Established Top-Performing Machine Learning Models}}
\label{table6}
\centering \includegraphics[width=8cm, height=13cm]{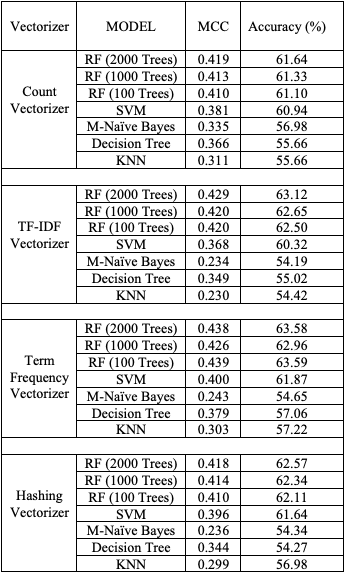}
\end{table}

Table \ref{table6} presents the results achieved by the established top-performing machine learning classifiers. Random Forest was consistently the top-performing model, out of all the five models, for all four vectorizer. \newline Random Forest with the Term-Frequency Vectorizer produced the highest accuracy of \textbf{63.59\%} and highest MCC = \textbf{0.438}. It was followed closely by SVM with Term-Frequency, which delivered an accuracy of \textbf{61.87\%} and MCC = \textbf{0.400}. Results obtained from the Term-Frequency Vectorizer were superior out of all four vectorizers in consideration. Count Vectorizer produced the worst result across all models. Multinomial-Naïve Bayes, Decision Tree and K-Nearest Neighbors were the poor performing models, consistently delivering an accuracy value below \textbf{60\%}, across all four vectorizers. A similar order was observed for the top-performing models i.e. Random Forest and SVM, where both of the models consistently performed better than Multinomial-Naïve Bayes, Decision Tree and K-Nearest Neighbors, delivering an accuracy above \textbf{60\%} across all four vectorizers.
To ensure that the values of Accuracy and MCC for different classifiers were not obtained by chance, the Friedman statistical test \cite{friedman1940} was performed. According to the Friedman test, the Null Hypothesis (H$^0$) states that there is no significant difference between the results achieved by the evaluated classification models. The Alternate Hypothesis (H$^1$) states that there is a significant difference between the performances achieved by the evaluated classifiers. The p-value obtained from the test was less than alpha = 0.05 (p-value\textsubscript{acc} = 0.015, p-value\textsubscript{mcc} = 0.023), which means that the Null Hypothesis (H0) can be rejected with a 95\% confidence level and the results obtained from the classifiers were statistically different.

Additionally, the ROC (Receiver Operating Characteristic) Curve and the AUC (Area Under Curve) was computed for the top-performing model (i.e. XLM) and the top-performing classification model (i.e. Random-Forest Classifier with Term Frequency Vectorizer). The results obtained from the corresponding curves validated the results obtained in the form of accuracy and MCC values. Through the AUC obtained for the respective labels, it was evident that XLM outperformed Random Forest (with Term-Frequency Vectorizer) in correctly predicting all the labels. XLM performed the best in correctly predicting Label 1 with an accuracy of 98\% and the worst in predicting Label 4 with an accuracy of 79\%. Random Forest (with Term-Frequency Vectorizer) displayed a similar order of accuracy for the respective labels with a value of 92\% for label 1 and 72\% for label 4. The high initial value of Recall (Sensitivity) for all the labels means a high true positive rate which corresponds to a trade-off with the Precision value. Figure \ref{figure5} and Figure \ref{figure6} show the ROC curve and AUC obtained for the top-performing BERT model and traditional classification model respectively.

\section{Discussions}

This section succinctly discusses the limitations, challenges and findings of this study. There is a large audience within the Indian YouTube community which uses Malayalam-English mix-code as its preferred medium for communication. While text classification using traditional classification models have shown promising results on monolingual datasets, the lack of optimization and support for this mix-code by both BERT and traditional classification models presented its own set of challenges.

Stemming and stopword removal are crucial pre-processing steps in NLP which can potentially hamper the performance of a model. While existing pre-processing libraries have been effective in processing monolingual datasets, this study has shown that standard stemming and stopword removal techniques need to be optimized for multilingual datasets. Therefore, Malayalam words were not stemmed and accordingly, stopwords written in the Malayalam script were not filtered out. Additionally, the lack of reputed sources for Mix-Code datasets, such as Malayalam-English presented another challenge which affects both current and future studies in this field of research. The available sources such as YouTube and other forms of social media are highly informal and riddled with spam, slang, spelling errors, abbreviations, emoticons and special characters. The aforementioned reasons could have potentially impacted the performance of the models.

A study presented by Chakravarthi \textit{et.} al. \cite{chakravarthi2020sentiment} also worked towards better understanding of Malayalam-English Mix-Code datasets. On comparison it was observed that our study extended the research by using a more diverse set of feature engineering techniques in the form of 4 Vectorizers. This provided insight into the performance of traditional models when trained with different feature engineering techniques. Further, in contrast to Chakravarthi \textit{et.} al. \cite{chakravarthi2020sentiment}, our study paid special attention to hyper-parameter tuning for both BERT as well as traditional classification models. Another study \cite{6968548} in the Malayalam-English Mix-Code domain proposed a rule based approach to analyse the sentiment of Malayalam movie reviews. In contrast to the rule based approach, where simply the count of negative, positive and neutral tokens is taken, applying Machine Learning is a more pragmatic approach with the scope of reproducibility. Moreover, the lack of feature engineering in Nait \textit{et.} al. \cite{6968548} is accounted for in our study, which is highly imperative while performing NLP tasks such as Text Classification.

The comparative study between the advanced language models and traditional classification models brought forth an insightful trade-off. Although, the transformer based language models were the more consistent and better performing models, they were computationally more expensive in terms of both time and space. This can be potentially due to the more extensive and robust design of the multi-layered Neural Network that is the heart of the language models. This trade-off can prove to be insightful for academic and research purposes as this breakthrough in NLP by the BERT language model can improve and build on existing research work. However, at an industry-level, established top-performing models might still be preferred due to their reliable yet efficient performance.

\section{Conclusion}
With the advent of YouTube as a video content publication platform with its monetization model and social networking features, its popularity has influenced a lot of content creators to broadcast their content via this medium. Viewer feedback in the form of comments is the sole channel of interaction with the audience for a content creator. But due to high volume, going through each and every comment is virtually impossible. Given this scenario, the primary objective of this study was to find promising techniques and settings that could be used for comment classification that could help creators filter comments to gain more insight on viewer response and in-turn improve future content.

The results have indicated that all the transformer based language models have consistently performed better than established classification models. In fact, the top 3 configurations of all 3 language models were able to achieve accuracy rates between 65\%-67\% (\textbf{XLM achieved the highest accuracy rate of 67.31\%}), while the most accurate classification model, i.e. Random Forest (2000 Trees) with Term-Frequency Vectorizer, could achieve an accuracy rate of 63.58\%. Moreover, performing the Friedman test eliminated the possibility of achieving the results by chance. Hence, we could come to a conclusion that the novel transformer based language models have proven to be better multi-text classifiers than established classification models for the given dataset. 

The fact that multiple models achieved similar results shows that there is scope in the future to use an ensemble of models to achieve better results. Additionally, supervised algorithms can learn from similar studies to better handle spam, slang, emoticons and special characters. BERT being a novel language model with limited support for languages other than English, shows that there is a lot of scope of improvement in that aspect to better process multilingual datasets. Subsequently, BERT can be used in conjunction with other NLP techniques such as Word Embedding, as done in other studies of emotion classification. The proposed methodologies in these studies can be extended to handle comment categorization. 

\section{Acknowledgement}
This research received no external funding.
\printbibliography

@inproceedings{sornsoontorn2017using,
  title={Using Document Classification to Improve the Performance of a Plagiarism Checker},
  author={Sornsoontorn, Chanchana and Rimcharoen, Sunisa and Leelathakul, Nutthanon and Kawtrakul, Asanee and Ratanaworabhan, Paruj},
  booktitle={2017 21st International Computer Science and Engineering Conference (ICSEC)},
  pages={1--5},
  year={2017},
  organization={IEEE}
}

@inproceedings{aman2007identifying,
  title={Identifying expressions of emotion in text},
  author={Aman, Saima and Szpakowicz, Stan},
  booktitle={International Conference on Text, Speech and Dialogue},
  pages={196--205},
  year={2007},
  organization={Springer}
}

@Report{Anthon2017text,
  Title                    = {Text classification of short messages (Detecting inappropriate comments in online user debates)},
  Author                   = {Anton Lundborg},
  Date                     = {2017-08-01},
  Institution              = {Lund University},
  Type                     = {Report},
  Year                     = {2017},
  Location                 = {Lund, Sweden},
  Version                  = {2017-08-01}
}

@inproceedings{durant2006mining,
  title={Mining sentiment classification from political web logs},
  author={Durant, Kathleen T and Smith, Michael D},
  booktitle={Proceedings of Workshop on Web Mining and Web Usage Analysis of the 12th ACM SIGKDD International Conference on Knowledge Discovery and Data Mining (WebKDD-2006), Philadelphia, PA},
  year={2006}
}

@inproceedings{wang2003classification,
  title={Classification of web documents using a naive bayes method},
  author={Wang, Yong and Hodges, Julia and Tang, Bo},
  booktitle={Proceedings. 15th IEEE International Conference on Tools with Artificial Intelligence},
  pages={560--564},
  year={2003},
  organization={IEEE}
}

@article{kalyani2016stock,
  title={Stock trend prediction using news sentiment analysis},
  author={Kalyani, Joshi and Bharathi, Prof and Jyothi, Prof and others},
  journal={arXiv preprint arXiv:1607.01958},
  year={2016}
}

@article{anne2017advanced,
  title={Advanced Text Analytics and Machine Learning Approach for Document Classification},
  author={Anne, Chaitanya},
  year={2017}
}

@mastersthesis{kleverwal2015supervised,
  title={Supervised text classification of medical triage reports},
  author={Kleverwal, Jurgen},
  year={2015},
  school={University of Twente}
}

@article{crijns2016text,
  title={Text Classification-Classifying events to ugenda calendar genres},
  author={Crijns, T},
  year={2016}
}

@Report{tobias2017automatic,
Title                    = {Automatic web page categorization using text classification methods},
Author                   = {Tobias Eriksson},
Date                     = {2013-08-08},
Institution              = {KTH Ventenskap},
Type                     = {Report},
Year                     = {2013},
Location                 = {Stockholm, Sweden},
Version                  = {2013-08-08}
}

@inproceedings{heidarysafa2018analysis,
  title={Analysis of Railway Accidents' Narratives Using Deep Learning},
  author={Heidarysafa, Mojtaba and Kowsari, Kamran and Barnes, Laura and Brown, Donald},
  booktitle={2018 17th IEEE International Conference on Machine Learning and Applications (ICMLA)},
  pages={1446--1453},
  year={2018},
  organization={IEEE}
}

@inproceedings{li2017text,
  title={Text classification method based on convolution neural network},
  author={Li, Lin and Xiao, Linlong and Wang, Nanzhi and Yang, Guocai and Zhang, Jianwu},
  booktitle={2017 3rd IEEE International Conference on Computer and Communications (ICCC)},
  pages={1985--1989},
  year={2017},
  organization={IEEE}
}

@inproceedings{grawe2017automated,
  title={Automated patent classification using word embedding},
  author={Grawe, Mattyws F and Martins, Claudia A and Bonfante, Andreia G},
  booktitle={2017 16th IEEE International Conference on Machine Learning and Applications (ICMLA)},
  pages={408--411},
  year={2017},
  organization={IEEE}
}

@inproceedings{shu2018investigating,
  title={Investigating Lstm with k-Max Pooling for Text Classification},
  author={Shu, Bo and Ren, Fuji and Bao, Yanwei},
  booktitle={2018 11th International Conference on Intelligent Computation Technology and Automation (ICICTA)},
  pages={31--34},
  year={2018},
  organization={IEEE}
}

@inproceedings{zharmagambetov2015sentiment,
  title={Sentiment analysis of a document using deep learning approach and decision trees},
  author={Zharmagambetov, Arman S and Pak, Alexandr A},
  booktitle={2015 Twelve international conference on electronics computer and computation (ICECCO)},
  pages={1--4},
  year={2015},
  organization={IEEE}
}

@inproceedings{zhao2017commented,
  title={Commented content classification with deep neural network based on attention mechanism},
  author={Zhao, Qinlu and Cai, Xiaodong and Chen, Chaocun and Lv, Lu and Chen, Mingyao},
  booktitle={2017 IEEE 2nd Advanced Information Technology, Electronic and Automation Control Conference (IAEAC)},
  pages={2016--2019},
  year={2017},
  organization={IEEE}
}

@inproceedings{kundu2018classification,
  title={Classification of Short-Texts Generated During Disasters: A Deep Neural Network Based Approach},
  author={Kundu, Shamik and Srijith, PK and Desarkar, Maunendra Sankar},
  booktitle={2018 IEEE/ACM International Conference on Advances in Social Networks Analysis and Mining (ASONAM)},
  pages={790--793},
  year={2018},
  organization={IEEE}
}

@inproceedings{chen2017deep,
  title={A deep-learning based ultrasound text classifier for predicting benign and malignant thyroid nodules},
  author={Chen, Dehua and Niu, Jinxuan and Pan, Qiao and Li, Yue and Wang, Mei},
  booktitle={2017 International Conference on Green Informatics (ICGI)},
  pages={199--204},
  year={2017},
  organization={IEEE}
}

@inproceedings{lakhotia2018experimental,
  title={An Experimental Comparison of Text Classification Techniques},
  author={Lakhotia, Suyash and Bresson, Xavier},
  booktitle={2018 International Conference on Cyberworlds (CW)},
  pages={58--65},
  year={2018},
  organization={IEEE}
}

@inproceedings{afakh2017aksara,
  title={Aksara jawa text detection in scene images using convolutional neural network},
  author={Afakh, Muhammad Labiyb and Risnumawan, Anhar and Anggraeni, Martianda Erste and Tamara, Mohamad Nasyir and Ningrum, Endah Suryawati},
  booktitle={2017 International Electronics Symposium on Knowledge Creation and Intelligent Computing (IES-KCIC)},
  pages={77--82},
  year={2017},
  organization={IEEE}
}

@inproceedings{behzadi2018text,
  title={Text Detection in Natural Scenes using Fully Convolutional DenseNets},
  author={Behzadi, Mitra and Safabakhsh, Reza},
  booktitle={2018 4th Iranian Conference on Signal Processing and Intelligent Systems (ICSPIS)},
  pages={11--14},
  year={2018},
  organization={IEEE}
}

@inproceedings{ameen2018spam,
  title={Spam detection in online social networks by deep learning},
  author={Ameen, Aso Khaleel and Kaya, Buket},
  booktitle={2018 International Conference on Artificial Intelligence and Data Processing (IDAP)},
  pages={1--4},
  year={2018},
  organization={IEEE}
}

@inproceedings{li2018electronic,
  title={Electronic Medical Data Analysis Based on Word Vector and Deep Learning Model},
  author={Li, Lan-Juan and Niu, Chao-Qun and Pu, Dong-Xu and Jin, Xin-Yu},
  booktitle={2018 9th International Conference on Information Technology in Medicine and Education (ITME)},
  pages={484--487},
  year={2018},
  organization={IEEE}
}

@inproceedings{uysal2017sentiment,
  title={Sentiment classification: Feature selection based approaches versus deep learning},
  author={Uysal, Alper Kursat and Murphey, Yi Lu},
  booktitle={2017 IEEE International Conference on Computer and Information Technology (CIT)},
  pages={23--30},
  year={2017},
  organization={IEEE}
}

@inproceedings{thazhackal2018hybrid,
  title={A Hybrid Deep Learning Model to Predict Business Closure from Reviews and User Attributes Using Sentiment Aligned Topic Model},
  author={Thazhackal, Sharun S and Devi, V Susheela},
  booktitle={2018 IEEE Symposium Series on Computational Intelligence (SSCI)},
  pages={397--404},
  year={2018},
  organization={IEEE}
}

@inproceedings{phaisangittisagul2019target,
  title={Target Advertising Classification using Combination of Deep Learning and Text model},
  author={Phaisangittisagul, E and Koobkrabee, Y and Wirojborisuth, K and Ratanasrimetha, T and Aummaro, S},
  booktitle={2019 10th International Conference of Information and Communication Technology for Embedded Systems (IC-ICTES)},
  pages={1--4},
  year={2019},
  organization={IEEE}
}

@article{madisetty2018neural,
  title={A neural network-based ensemble approach for spam detection in Twitter},
  author={Madisetty, Sreekanth and Desarkar, Maunendra Sankar},
  journal={IEEE Transactions on Computational Social Systems},
  volume={5},
  number={4},
  pages={973--984},
  year={2018},
  publisher={IEEE}
}

@inproceedings{ray2015text,
  title={Text recognition using deep blstm networks},
  author={Ray, Anupama and Rajeswar, Sai and Chaudhury, Santanu},
  booktitle={2015 eighth international conference on advances in pattern recognition (ICAPR)},
  pages={1--6},
  year={2015},
  organization={IEEE}
}

@inproceedings{wang2019research,
  title={Research on hot news classification algorithm based on deep learning},
  author={Wang, Zaiying and Song, Bohao},
  booktitle={2019 IEEE 3rd Information Technology, Networking, Electronic and Automation Control Conference (ITNEC)},
  pages={2376--2380},
  year={2019},
  organization={IEEE}
}

@inproceedings{chen2018hybrid,
  title={A Hybrid Deep Learning Model for Text Classification},
  author={Chen, Xianglong and Ouyang, Chunping and Liu, Yongbin and Luo, Lingyun and Yang, Xiaohua},
  booktitle={2018 14th International Conference on Semantics, Knowledge and Grids (SKG)},
  pages={46--52},
  year={2018},
  organization={IEEE}
}

@article{kaushik2016comprehensive,
  title={A comprehensive study of text mining approach},
  author={Kaushik, Abhishek and Naithani, Sudhanshu},
  journal={International Journal of Computer Science and Network Security (IJCSNS)},
  volume={16},
  number={2},
  pages={69},
  year={2016},
  publisher={International Journal of Computer Science and Network Security}
}

@article{kaur2019cooking,
  title={Cooking Is Creating Emotion: A Study on Hinglish Sentiments of Youtube Cookery Channels Using Semi-Supervised Approach},
  author={Kaur, Gagandeep and Kaushik, Abhishek and Sharma, Shubham},
  journal={Big Data and Cognitive Computing},
  volume={3},
  number={3},
  pages={37},
  year={2019},
  publisher={Multidisciplinary Digital Publishing Institute}
}

@article{friedman1940,
author = "Friedman, Milton",
doi = "10.1214/aoms/1177731944",
fjournal = "Annals of Mathematical Statistics",
journal = "Ann. Math. Statist.",
month = "03",
number = "1",
pages = "86--92",
publisher = "The Institute of Mathematical Statistics",
title = "A Comparison of Alternative Tests of Significance for the Problem of $m$ Rankings",
url = "https://doi.org/10.1214/aoms/1177731944",
volume = "11",
year = "1940"
}

@dataset{kazhuparambil_subramaniam_2020_3871306,
  author={Kazhuparambil,  Subramaniam and
          Kaushik, Abhishek and
          Pichakassery Hari, Athira and
          Sanu, Vivek and
          Ramchandra Uppari, Rahul and
          YADAV, HARISH},
  title={{MALAYALAM LANGUAGE (MIX CODE) Recipe channels 
          Youtube Comments}},
  month=jun,
  year=2020,
  publisher={Zenodo},
  version={Version1},
  doi={10.5281/zenodo.3871306},
  url={https://doi.org/10.5281/zenodo.3871306}
}

@article{shah2020opinion,
  title={Opinion-Mining on Marglish and Devanagari Comments of YouTube Cookery Channels Using Parametric and Non-Parametric Learning Models},
  author={Shah, Sonali Rajesh and Kaushik, Abhishek and Sharma, Shubham and Shah, Janice},
  journal={Big Data and Cognitive Computing},
  volume={4},
  number={1},
  pages={3},
  year={2020},
  publisher={Multidisciplinary Digital Publishing Institute}
}

@article{8975793,
title={The Automatic Text Classification Method Based on BERT and Feature Union},
author={W. {Li} and S. {Gao} and H. {Zhou} and Z. {Huang} and K. {Zhang} and W. {Li}},  
booktitle={2019 IEEE 25th International Conference on Parallel and Distributed Systems (ICPADS)},      
year={2019},  
volume={},  
number={},  
pages={774-777}}

@article{8959920,
title={Japanese abstractive text summarization using BERT},
author={Y. {Iwasaki} and A. {Yamashita} and Y. {Konno} and K. {Matsubayashi}},  
booktitle={2019 International Conference on Technologies and Applications of Artiﬁcial Intelligence (TAAI)}, year={2019},  
volume={},  
number={},  
pages={1-5}
}

@article{shah2019sentiment,
  title={Sentiment Analysis On Indian Indigenous Languages: A Review On Multilingual Opinion Mining},
  author={Shah, Sonali Rajesh and Kaushik, Abhishek},
  journal={arXiv preprint arXiv:1911.12848},
  year={2019}
}

@incollection{venkatakrishnan2020sentiment,
  title={Sentiment Analysis on Google Play Store Data Using Deep Learning},
  author={Venkatakrishnan, Swathi and Kaushik, Abhishek and Verma, Jitendra Kumar},
  booktitle={Applications of Machine Learning},
  pages={15--30},
  year={2020},
  publisher={Springer}
}

@article{kaushikstudy,
  title={A Study on Sentiment Analysis: Methods and Tools},
  author={Kaushik, Abhishek and Kaushik, Anchal and Naithani, Sudhanshu}
}

@article{chakravarthi2020sentiment,
    title={A Sentiment Analysis Dataset for Code-Mixed Malayalam-English},
    author={Bharathi Raja Chakravarthi and Navya Jose and Shardul Suryawanshi and Elizabeth Sherly and John P. McCrae},
    year={2020},
    eprint={2006.00210},
    archivePrefix={arXiv},
    primaryClass={cs.CL}
}

@article{6968548,
  title={SentiMa - Sentiment extraction for Malayalam}, 
  author={D. S. {Nair} and J. P. {Jayan} and R. R. {R} and E. {Sherly}},
  booktitle={2014 International Conference on Advances in Computing, Communications and Informatics (ICACCI)}, 
  year={2014},
  volume={},
  number={},
  pages={1719-1723}
 }
%\bibliographystyle{IEEEtr}  
%\bibliography{references}

\end{document}